\let\mypdfximage\pdfximage
\def\pdfximage{\immediate\mypdfximage}
\documentclass[10pt,twocolumn,letterpaper]{article}

\usepackage{cvpr}              % To produce the CAMERA-READY version
\usepackage[accsupp]{axessibility}
\usepackage{graphicx}
\usepackage{amsmath}
\usepackage{amssymb}
\usepackage{booktabs}
\usepackage{todonotes}
\usepackage{hhline}
\usepackage{multirow}
\usepackage{adjustbox}

\usepackage[pagebackref,breaklinks,colorlinks]{hyperref}
\usepackage{breakurl}

\usepackage[capitalize]{cleveref}
\crefname{section}{Sec.}{Secs.}
\Crefname{section}{Section}{Sections}
\Crefname{table}{Table}{Tables}
\crefname{table}{Tab.}{Tabs.}

\def\cvprPaperID{} % *** Enter the CVPR Paper ID here
\def\confName{CVPR-WAD}
\def\confYear{2023}
\newcommand{\paperTitle}[0]{Does Image Anonymization Impact Computer Vision Training?}

\newcommand{\appExpDetailsCOCOInstance}[0]{Appendix \textcolor{red}{A.2}\xspace}
\newcommand{\appExpDetailsCityscapesNoPersons}[0]{Appendix \textcolor{red}{A.3}\xspace}
\newcommand{\appQualitative}[0]{Appendix \textcolor{red}{B}\xspace}

\begin{document}

%%%%%%%%% TITLE - PLEASE UPDATE
\title{\paperTitle}

\author{Håkon Hukkelås
\and
Frank Lindseth
\and
Deparment of Computer Science, Norwegian University of Science and Technology 
\and 
hakon.hukkkelas@ntnu.no
}

\twocolumn[{%
\renewcommand\twocolumn[1][]{#1}%
\maketitle
\begin{center}
    \captionsetup{type=figure}
    \includegraphics[width=.94\textwidth]{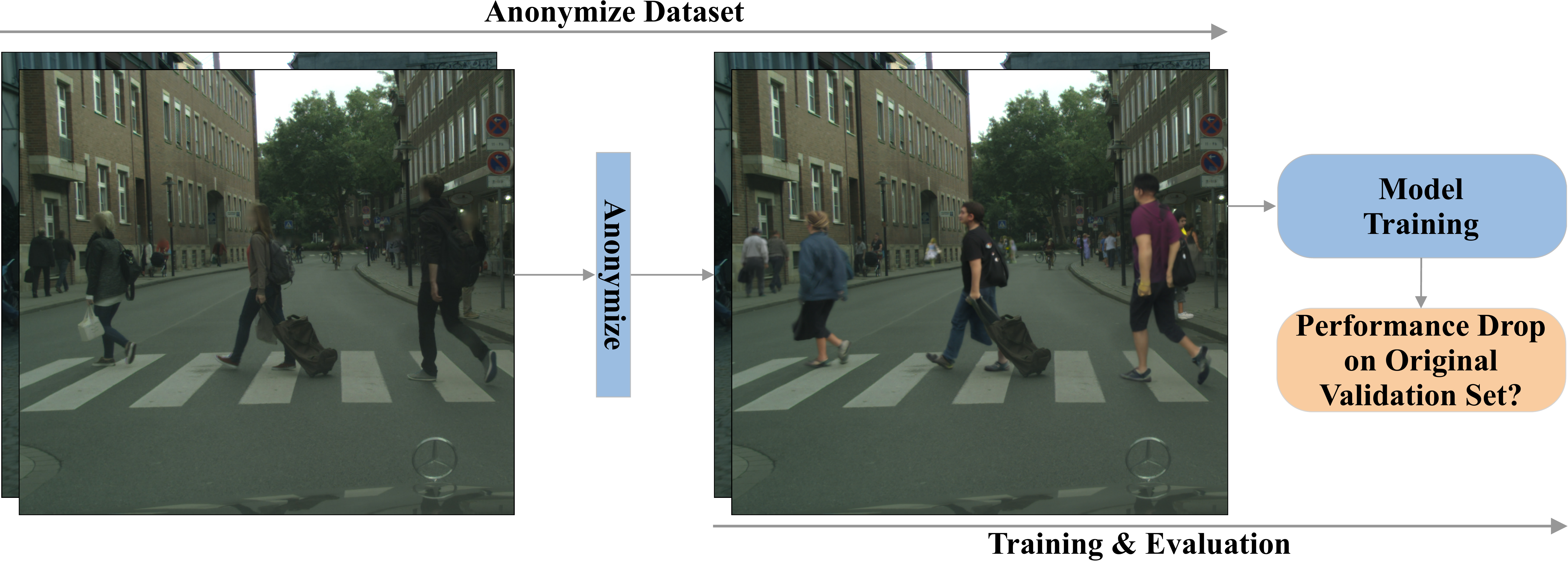}
    \captionof{figure}{
        To assess the impact of anonymization, we first anonymize common computer vision datasets, then train various models using the anonymized data, and finally evaluate the models on the original validation datasets.
        The figure depicts our Cityscapes \cite{cityscapes} full-body anonymization experiment.
        Note that the leftmost image is anonymized with face blurring, following Cityscapes \cite{cityscapes} terms of use.
        }
        \label{fig:task_figure}%
\end{center}%
}]

%%%%%%%%% ABSTRACT
\begin{abstract}
Image anonymization is widely adapted in practice to comply with privacy regulations in many regions.
However, anonymization often degrades the quality of the data, reducing its utility for computer vision development.
In this paper, we investigate the impact of image anonymization for training computer vision models on key computer vision tasks (detection, instance segmentation, and pose estimation).
Specifically, we benchmark the recognition drop on common detection datasets, where we evaluate both traditional and realistic anonymization for faces and full bodies.
Our comprehensive experiments reflect that traditional image anonymization substantially impacts final model performance, particularly when anonymizing the full body.
Furthermore, we find that realistic anonymization can mitigate this decrease in performance, where our experiments reflect a minimal performance drop for face anonymization.
Our study demonstrates that realistic anonymization can enable privacy-preserving computer vision development with minimal performance degradation across a range of important computer vision benchmarks.
\end{abstract}

\section{Introduction}

Collecting and storing large amounts of visual data is a fundamental task in developing robust and efficient computer vision algorithms.
However, this raises concerns regarding the individual's right to privacy, as visual data is rich in privacy-sensitive information, \eg persons, license plates, and street signs.
Recent privacy legislation (\eg GDPR \cite{GDPR} in the European Union) requires anonymization when collecting visual data or consent from individuals, which is often infeasible.
This can be viewed as a barrier to research and development, particularly for the data-dependent field of Autonomous Vehicle (AV) research.
To compensate for these restrictions, practitioners have adopted traditional image anonymization (\eg blurring) for collecting AV datasets \cite{Geyer2020,Caesar2020} and street view images \cite{Frome2009}.

Traditional image anonymization can protect privacy, but it severely distorts the visual data, potentially reducing its utility for computer vision development.
Despite this, face obfuscation (\eg blurring) is the standard method employed to anonymize public autonomous vehicle datasets \cite{Geyer2020,Caesar2020}, and its impact on final model performance is currently unclear.
Previous work analyzed the impact of face anonymization for  classification \cite{Yang2022a}, semantic segmentation \cite{Geyer2020,Zhou2022}, object detection \cite{Dvoracek2022}, action recognition \cite{Tomei2021}, and face detection \cite{Klomp2021}.
In summary, their findings reveal that face anonymization can impact visual recognition related to the human class, and it can severely hurt tasks where the human is in focus \cite{Klomp2021,Tomei2021}.

Our literature review, detailed in \cref{sec:related_work}, resulted in two unanswered questions, which we address in this study.

First, \emph{is realistic anonymization more effective to preserve image utility compared to traditional methods?}
Realistic anonymization replaces privacy-sensitive information with synthesized content from generative models, which are found to better preserve utility compared to traditional methods \cite{Sun2018,Hukkelas2019}.
Previous work has found realistic anonymization to improve utility preservation for semantic segmentation \cite{Klomp2021,Zhou2022}.
Our work builds upon this by investigating different objectives and datasets.

Secondly, \emph{to what extent does full-body anonymization impact the training of computer vision models?}
The human body is recognizable from many cues outside the face (\eg gait, clothes, ear, body shape), often requiring full-body anonymization to protect privacy.
A few studies explore the impact of full-body anonymization \cite{Hukkelas2022,Hukkelas2022a}, where they find it to improve over traditional methods.
However, they rely on automatic detection methods, which opens the question if the performance degradation is due to detection errors or the anonymization model.
Furthermore, their model requires dense pose estimation \cite{AlpGuler2018,Neverova2020}, which limits anonymization to individuals close to the camera due to limited long-range detection recall of dense pose models.

In this paper, we focus on key computer vision tasks related to autonomous vehicles, namely instance segmentation and human pose estimation.
We evaluate the full-body and face anonymization models built in DeepPrivacy2 \cite{Hukkelas2022a} and compare realistic anonymization to traditional methods.
See \burl{https://github.com/hukkelas/deep_privacy2/blob/master/docs/anonymizing_datasets.md} to reproduce our experiments.

\begin{figure*}
    \begin{subfigure}{0.166666666666666666666666667\textwidth}
        \includegraphics[width=\textwidth,trim={0 5cm 0 5cm},clip]{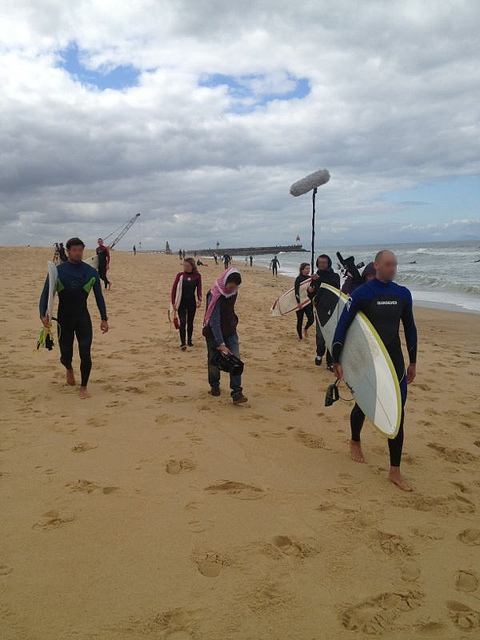}
        \caption{Face - Gaussian}
    \end{subfigure}%
    \begin{subfigure}{0.166666666666666666666666667\textwidth}
        \includegraphics[width=\textwidth,trim={0 5cm 0 5cm},clip]{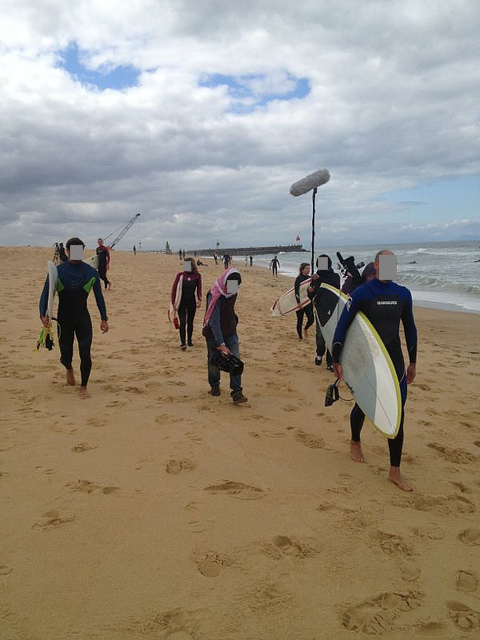}
        \caption{Face - Maskout}
    \end{subfigure}%
    \begin{subfigure}{0.166666666666666666666666667\textwidth}
        \includegraphics[width=\textwidth,trim={0 5cm 0 5cm},clip]{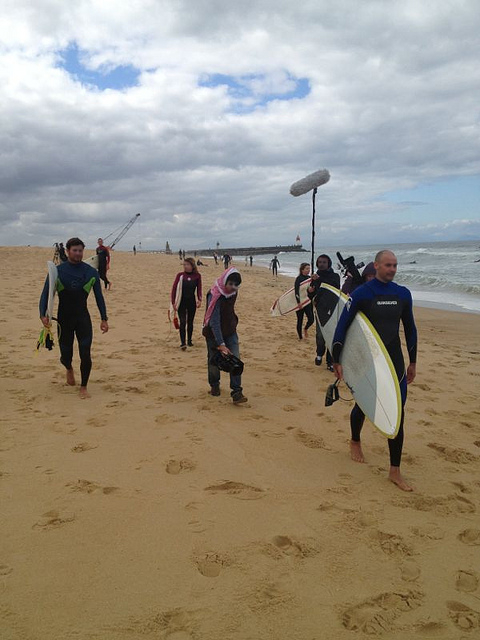}
        \caption{Face - Realistic}
    \end{subfigure}%
    \begin{subfigure}{0.166666666666666666666666667\textwidth}
        \includegraphics[width=\textwidth,trim={0 5cm 0 5cm},clip]{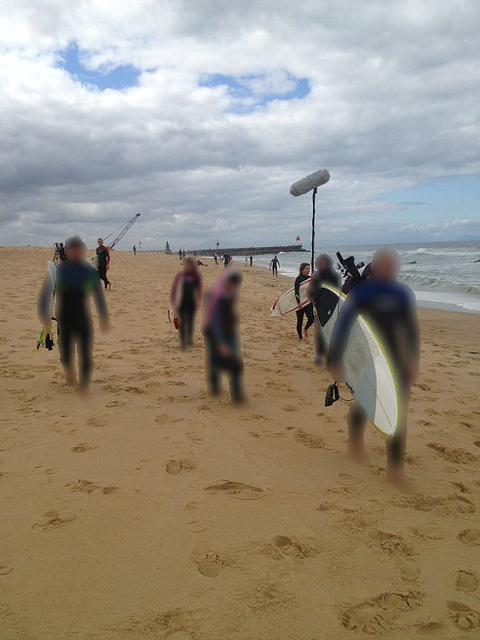}
        \caption{Body - Gaussian}
    \end{subfigure}%
    \begin{subfigure}{0.166666666666666666666666667\textwidth}
        \includegraphics[width=\textwidth,trim={0 5cm 0 5cm},clip]{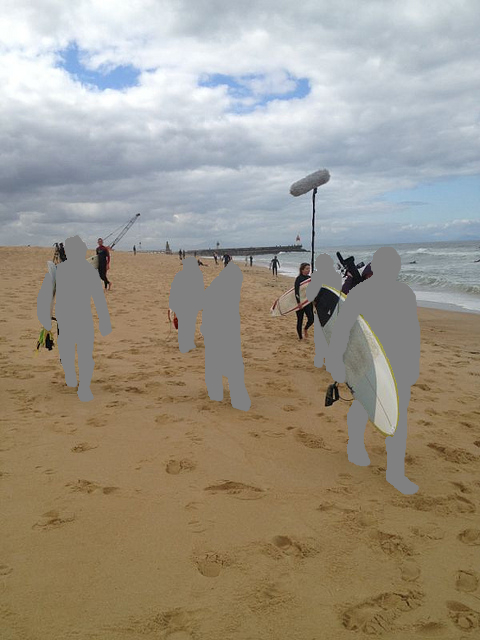}
        \caption{Body - Mask out}
    \end{subfigure}%
    \begin{subfigure}{0.166666666666666666666666667\textwidth}
        \includegraphics[width=\textwidth,trim={0 5cm 0 5cm},clip]{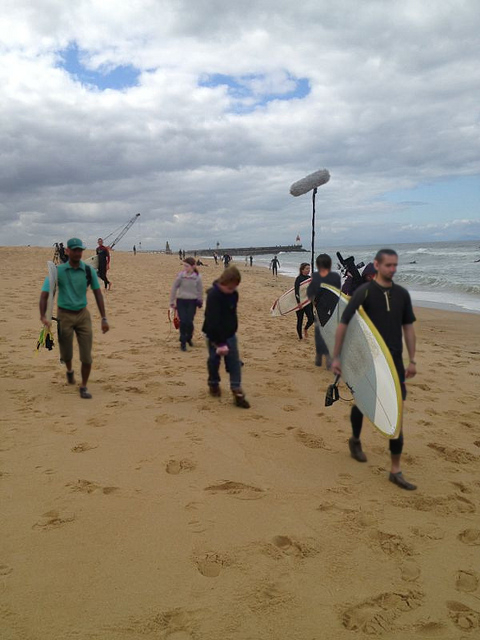}
        \caption{Body - Realistic}
    \end{subfigure}%
    \caption{The different anonymization methods evaluated in this paper. Image from COCO train2017 \cite{Lin2014COCO}, image id=000000097507.}
    \label{fig:anonymization_comparison}
\end{figure*}
\section{Related Work}
\label{sec:related_work}
\paragraph{Image Anonymization}
The goal of image anonymization is to remove any privacy-sensitive information contained in the image.
Traditional anonymization is widely adopted in practice, where methods anonymize the image via obfuscation (\eg blurring, masking), encryption \cite{He2016Puppies}, or k-means \cite{Gross2006a,Jourabloo2015,Newton2005}.
Often, these methods are sufficient to protect privacy; however, they degrade the quality of the data reducing its utility for downstream tasks.

Recent work has introduced \emph{realistic image anonymization}, where anonymization is done by replacing persons with synthesized identities from a  generative model.
The majority of previous work focuses on face anonymization, where current methods anonymize by \emph{inpainting} a masked out region \cite{Hukkelas2019,Maximov2020,Sun2018,Sun2018a}, or \emph{transforming} \cite{Gafni2019,Ren2018a,Ciftci2023MFMC} the original identity to remove privacy-sensitive information.
Transformative models often maintain higher utility (\eg preserving facial expression) but offer no formal guarantee of removing the original identity  from the image, making them vulnerable to adversarial attacks.
A few methods explore anonymizing the full-body \cite{Hukkelas2022a,Hukkelas2022,Brkic2017,Maximov2020}, where the current state-of-the-art \cite{Hukkelas2022a,hukkelas23synthesizing} can generate convincing full-bodies given sparse keypoints \cite{hukkelas23synthesizing} or dense pose annotations \cite{Hukkelas2022a}.
Finally, some methods insert adversarial perturbation in the image, which is invisible to the human eye but able to fool face recognition systems \cite{Oh2017}.

\paragraph{Privacy Guarantees of Anonymization}
Most current anonymization systems offer no formal guarantee of anonymization, and the identity can often be recognized from other cues in the image.
Image blurring is discussed numerous times in the literature \cite{Li2017c,Li2017b,Boyle2000,Neustaedter2006,Gross2009,Newton2005}, where the identity is often recognizable due to limited blurring.
Furthermore, the identity is recognizable even though the face is anonymized through other identifying attributes of the human body \cite{Wilber2016,Lander01,McPherson2016}, such as gait \cite{Abarca2021handbook}, clothing \cite{Gallagher2008}, and body appearance \cite{Zhang2015,Oh2016}.
This makes full-body anonymization more effective than face anonymization in terms of privacy.
Finally, most anonymization systems rely on automatic detection, which is far from perfect and vulnerable to adversarial attacks \cite{Kurakin2019AdversarialExamples}. 

\paragraph{Public Anonymized Datasets}
The prominent computer vision datasets employ no form of anonymization, where only a few datasets are anonymized.
NuScenes \cite{Caesar2020} contains images from vehicles driving in Singapore and Boston, where faces and license plates are anonymized via blurring.
A2D2 \cite{Geyer2020} includes data from southern Germany, where license plates and heads are blurred to comply with German privacy regulations.
AViD \cite{Piergiovanni2020} is a video dataset for action recognition with blurred heads.
P3M \cite{Li2021} is a portrait matting dataset where every face is blurred.
\cite{Uittenbogaard2019} propose a dataset containing street view scenes where cars and pedestrians are removed via image inpainting.

\paragraph{Visual Recognition on Anonymized Data}
There exists a limited set of studies exploring the effect that anonymization has on training computer vision models.
For ImageNet \cite{Deng2010ImageNet} training, face obfuscation (blurring) has little effect on top-5 accuracy and  no impact on feature transferability to scene recognition, object localization, and face attribute classification.
Nevertheless, anonymization slightly degrades accuracy in  classes appearing together with faces (\eg facial masks).
For autonomous vehicle datasets, traditional face anonymization can degrade instance segmentation on Cityscapes \cite{cityscapes,Zhou2022}, whereas realistic face anonymization has no noticeable negative impact.
Furthermore, they find that larger backbones and multi-scale features are more robust to image anonymization \cite{Zhou2022}.
Dvoracek \etal \cite{Dvoracek2022} finds little impact of face anonymization on object detection on the same dataset.
Geyer \cite{Geyer2020} finds that face anonymization has little effect on semantic segmentation on the A2D2 dataset.
For face detection, realistic anonymization performs substantially better than traditional methods for training face detectors \cite{Klomp2021}.
For action recognition, face obfuscation significantly degrades  performance \cite{Tomei2021}, where the authors propose a teacher-student self-distillation framework to mitigate the degradation.

Finally, we note that some studies focus on the human perspective and investigate the effect of different anonymization techniques on the users' perceived experience \cite{Hasan2018,Li2017c}.

\begin{figure*}
    \includegraphics[width=\textwidth]{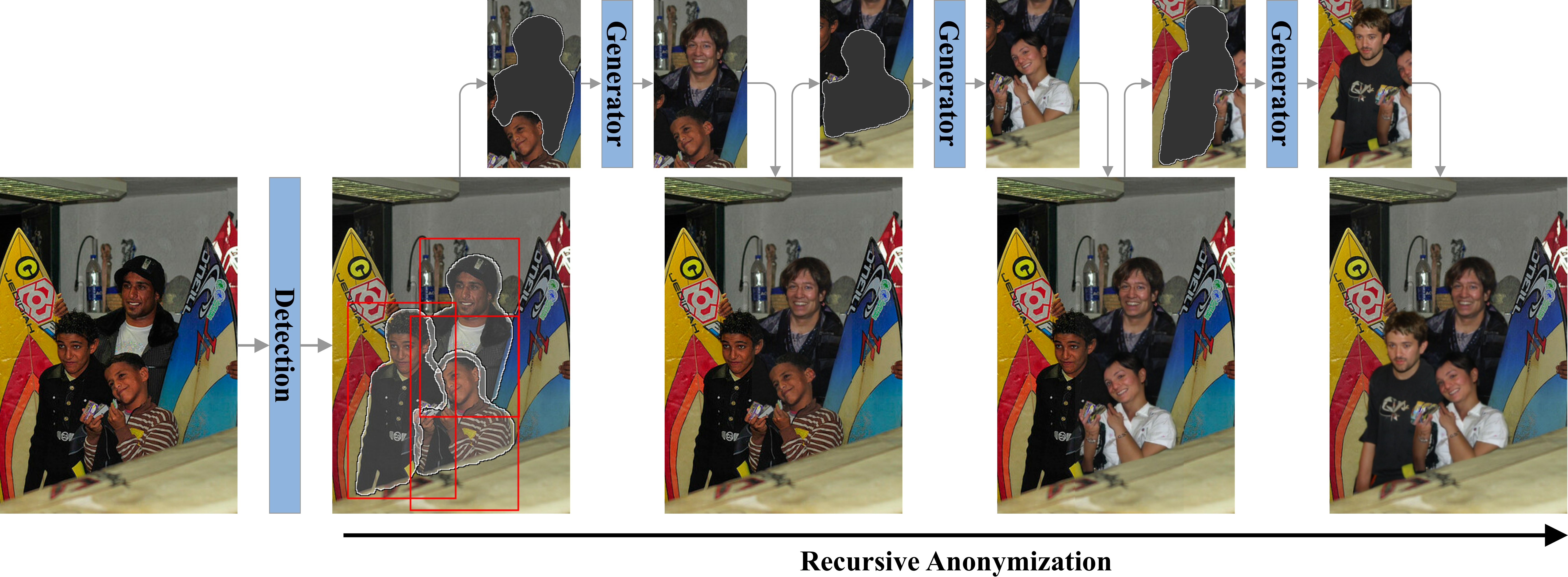}
    \caption{
        DeepPrivacy2 \cite{Hukkelas2022a} anonymizes one instance at a time, then paste each synthesized individual into the original image.
        For our experiments, detection is not performed, as segmentation masks are defined from pre-defined annotations (see \cref{sec:anonymization_region}).
        Note that the generator relies on keypoint annotations, which are not depicted here.
        }
    \label{fig:anonymization_pipeline}
\end{figure*}
\section{Anonymization Method}
\label{sec:method}

In this paper, we explore three different anonymization techniques for full-body and face anonymization; blurring, mask-out, and realistic anonymization (see \cref{fig:anonymization_comparison}).
Given the image $I$ and a mask $M$ indicating the region to be anonymized, the goal of each method is to remove any privacy-sensitive information within $M$.
In this section, we first define $M$ for face and full-body anonymization (\Cref{sec:anonymization_region}), then introduce the anonymization methods in \Cref{sec:traditional_method} and \Cref{sec:realistic_method}.

\subsection{Anonymization Region}
\label{sec:anonymization_region}
To define the anonymization region, we employ the pre-defined instance segmentation annotations for the person/pedestrian class, as every dataset in this paper includes such annotations.
Note that we do not anonymize annotations marked as "crowd" or "ignored" in the datasets, nor classes that often contain a person (\eg bicycle, motorcycle), as the realistic anonymization techniques require distinct instance-wise annotations.
Given the two aforementioned filtering criteria, it is important to note that  we are not able to anonymize all individuals in the dataset.
An alternative option is to obtain instance-wise annotations by manual annotation or automatic detection.
However, we decided against this approach, as the former is too time-consuming, and the latter may introduce detection errors, making it unclear if performance degradation is due to detection errors or poor anonymization.

\paragraph{Face Region}
As none of the benchmark datasets include annotated faces, we define the face anonymization region following a standard face detection dataset, WIDER-Face \cite{Yang2016WIDER}.
Specifically, the region is the minimal bounding box containing the forehead, chin, and cheek.
We annotate each dataset with a pre-trained face detector (DSFD \cite{Li2018c}), where we filter the detections by matching them with annotated instance segmentations.
We match boxes to segmentations via Intersection over Union (IoU), where we select the match with the highest IoU and bounding box score.
Any matches with an IoU $< 1\%$ are removed.

\paragraph{Full-Body Anonymization}
Since all benchmark datasets include annotated instance segmentations, we use these to define the full-body anonymization region.
To compensate for annotations where the segmentations don't fully encompass the body (often segmentation does not include bordering pixels), we slightly dilate the segmentation following \cite{Hukkelas2022a}.

\subsection{Traditional Anonymization}
\label{sec:traditional_method}
We evaluate two commonly used obfuscation techniques for traditional anonymization, namely blurring and masking out.
Note that we employ the same method for both face and full-body anonymization.

\paragraph{Mask-Out}
Mask-out defines the anonymized image as $I_{new} = I \odot (1-M) + M \odot 127$, where $\odot$ is element-wise multiplication.

\paragraph{Gaussian Blur}
Gaussian blur defines the anonymized image as $I_{new} = I \odot (1-M) + M \odot I_{blur}$.
Here, $I_{blur}$ is the blurred image with a Gaussian filter ($\sigma=7$, k-size= $3 \cdot \sigma$).

\subsection{Realistic Anonymization}
\label{sec:realistic_method}

For realistic anonymization, we employ pre-trained models from DeepPrivacy2 \cite{Hukkelas2022a}.
Note that DeepPrivacy2 anonymizes by inpainting (illustrated in \cref{fig:anonymization_pipeline}), such that it never observes the masked region in $I$.
Thus, it provides similar privacy protection as mask-out anonymization.

\paragraph{Face Anonymization}
For face anonymization, we employ the face anonymization model in DeepPrivacy2 \cite{Hukkelas2022a}, which is a U-Net GAN trained on FDF \cite{Hukkelas2019} that synthesizes faces at $128 \times 128$ resolution.
This model does not rely on keypoint annotations, which enables it to anonymize all faces detected.

\paragraph{Full-Body Anonymization}
For full-body anonymization, we employ a U-Net GAN \cite{hukkelas23synthesizing} relying on keypoint annotations following the COCO format \cite{Lin2014COCO}.
This model is trained on the FDH dataset \cite{Hukkelas2022a}, and the model is integrated into the DeepPrivacy2 framework \cite{Hukkelas2022a}.
For datasets without keypoint annotations, we use a top-down pose estimation network (ViTPose \cite{Xu22VITPose}) which estimates the pose given the image and the minimal bounding box encompassing the instance segmentation.
All keypoints with a confidence $\geq 30\%$ are assumed to be visible.

\subsection{Global Context for Full-Body Synthesis}
\label{sec:global_context}

\begin{figure*}
    \begin{subfigure}{0.14285714\textwidth}
        \includegraphics[width=\textwidth]{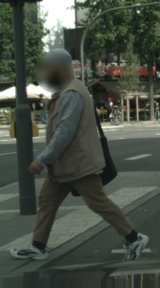}
        \caption*{Original}
    \end{subfigure}% 
    \begin{subfigure}{0.14285714\textwidth}
        \includegraphics[width=\textwidth]{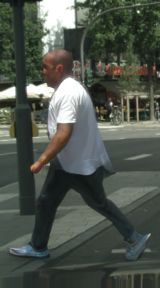}
        \caption*{Initial $\omega$}
    \end{subfigure}%
    \begin{subfigure}{0.14285714\textwidth}
        \includegraphics[width=\textwidth]{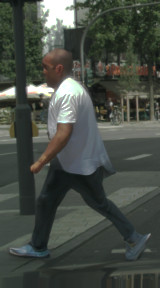}
        \caption*{HM}
    \end{subfigure}
    \begin{subfigure}{0.28571428\textwidth}
        \includegraphics[width=.5\textwidth]{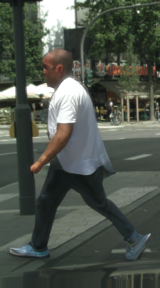}%
        \includegraphics[width=.5\textwidth]{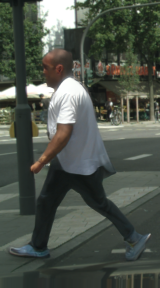}%
        \caption*{HM-LO Optimization $\rightarrow$ \hfil}
    \end{subfigure}%
    \begin{subfigure}{0.14285714\textwidth}
        \includegraphics[width=\textwidth]{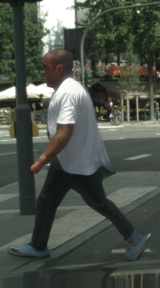}%
        \caption*{}
    \end{subfigure}%
    \begin{subfigure}{0.14285714\textwidth}
        \includegraphics[width=\textwidth]{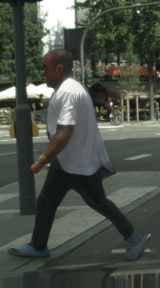}%
        \caption*{Final Image}
    \end{subfigure}%
    \caption{
    The initial synthesized identity ("initial $\omega$") may not align with the global context of the image, making the synthesized identity "stick out" compared to the original identity.
    We explore two options to address this issue: naive histogram matching (\textbf{HM}), and Histogram matching via latent optimization (\textbf{HM-LO}), which iteratively adjusts the initial $\omega$ to better fit the histogram of the original image (in HSV)
        }
    \label{fig:histogram_matching}
\end{figure*}

In our preliminary experiments, we observed that the full-body generative model often generated human bodies that fit the local context of the generative model but did not align with the global context.
We believe this is not a limitation of the generative model itself but a limitation to the crop-based anonymization method used by DeepPrivacy2 (see \cref{fig:anonymization_pipeline}).
In this paper, we explore two solutions to this issue; ad-hoc histogram equalization and histogram matching via latent optimization illustrated in \cref{fig:histogram_matching}

\begin{figure}
    \begin{subfigure}{0.165\textwidth}
        \includegraphics[width=\textwidth]{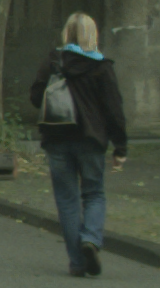}
        \caption*{Original}
    \end{subfigure}%
    \begin{subfigure}{0.165\textwidth}
        \includegraphics[width=\textwidth]{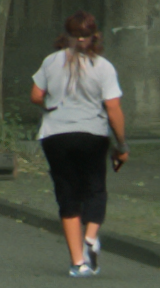}
        \caption*{Anonymized}
    \end{subfigure}%
    \begin{subfigure}{0.165\textwidth}
        \includegraphics[width=\textwidth]{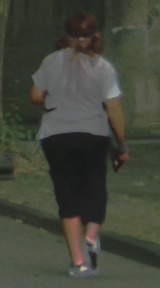}
        \caption*{Final after HM}
    \end{subfigure}
    \caption{
        Naive histogram matching can introduce visual artifacts.}
    \label{fig:failure_case_naive_histogram_matching}
\end{figure}

\paragraph{Histogram Matching (HM)}
A naive approach for matching the generated body to the global context is naive histogram equalization.
Specifically, we match the synthesized (cropped) image to the original (cropped) image by using skimage \href{https://scikit-image.org/docs/stable/api/skimage.exposure.html#skimage.exposure.match_histograms}{match\_histogram}.
This adjusts the synthesized image such that each color channel (RGB) matches the cumulative histogram of the original image.
To reduce bordering effects when pasting the equalized image into the original image, we smoothly transition the border by slightly blurring the mask with a gaussian filter.
That is, given the cropped image $x$, the corresponding mask $M_c$, and the synthesized image $y$, the new image is given by; $y_{new} = x \odot (1-M_c^{blurred}) + y \odot M_c^{blurred}$,
where $M_c^{blurred}$ is $M_c$ blurred with a gaussian filter with size=$[19, 19]$ and $\sigma=9$.
We note that this is far from an optimal solution, where naive histogram matching can introduce severe visual artifacts \cref{fig:failure_case_naive_histogram_matching}.

\paragraph{Histogram Matching via Latent Optimization (HM-LO)}
An alternative approach to post-processing the output is a search in the latent space of the generator.
Conceptually, if the exact environmental context (\eg scene lightning) is not given by the cropped image, it should be possible to adjust such factors through the latent space of the generator.
Therefore, we suggest utilizing gradient descent to modify the latent vector of the generator, aligning the histogram of the generated image with that of the original image

Given the cropped image $x$ and the mask $M_c$, the generated image is $y = G(x \odot M_C, \omega)$, where $\omega$ is the latent space of the generator, following StyleGAN \cite{Karras2018}.
Given $x$, we adjust a sampled $\omega$ via gradient descent such that $y$ matches the histogram of $x$ in the S and V channel of the HSV transform of $x$ and $y$. 
Specifically, we optimize;
\begin{equation}
    \begin{split}
    \mathcal{L}(x_{hsv}, y_{hsv}) = \mathbb{W}(P_S(x_{hsv}), P_S(y_{hsv})) + \\ 
    \mathbb{W}(P_V(x_{hsv}), P_V(y_{hsv})),
\end{split}
\end{equation}
where $\mathbb{W}$ is the Wasserstein-1 distance, and $P_V$, $P_S$ is the histogram of the S and V color channel in the HSV transformed image of $x$ and $y$.
Then, we perform gradient descent on $\omega$ for 100 steps or until $ \mathcal{L}(x_{hsv}, y_{hsv})< 0.02$.

\begin{figure}
    \begin{subfigure}{0.165\textwidth}
        \includegraphics[width=\textwidth]{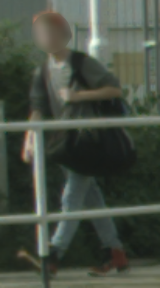}
        \caption*{Original}
    \end{subfigure}%
    \begin{subfigure}{0.165\textwidth}
        \includegraphics[width=\textwidth]{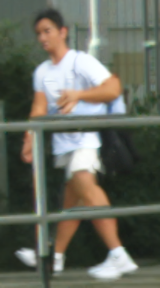}
        \caption*{Initial $\omega$}
    \end{subfigure}%
    \begin{subfigure}{0.165\textwidth}
        \includegraphics[width=\textwidth]{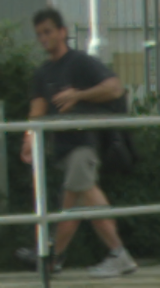}
        \caption*{Final $\omega$}
    \end{subfigure}
    \caption{
        Histogram Matching via Latent Optimization can induce significant semantic changes to the synthesized identity, due to directly optimizing $\omega$ to match the HSV histogram (S/V channels).}
    \label{fig:failure_case_histogram_matching}
\end{figure}

Often, HM-LO induces slight adjustments to the generated image such that it better matches the context of the image (\cref{fig:histogram_matching}).
However, we note that HM-LO can induce significant semantic changes if the original sampled colors deviate from the original identity (\cref{fig:failure_case_histogram_matching}).

\section{Experiments}
In this section, we report results for training on anonymized data.
We train each model on the anonymized dataset and report standard evaluation metrics on the original validation set.
To reduce randomness, we report the average and standard error over three independent training runs using seeds 0, 1, and 2.
All experiments are done with Pytorch 1.12 \cite{paszke2019pytorch} on a single NVIDIA A100-40GB.
Random qualitative examples from our experiments are given in \appQualitative.

\subsection{Experimental Details}

\paragraph{COCO Pose Estimation}
We train a Keypoint \href{https://github.com/facebookresearch/detectron2/blob/main/configs/COCO-Keypoints/keypoint_rcnn_R_50_FPN_3x.yaml}{R-50 FPN} R-CNN using detectron2 \cite{wu2019detectron2} on the COCO2017 dataset \cite{Lin2014COCO}.
The training dataset contains 118,287 images with 149,813 person instances (after filtration following \cref{sec:method}), and we evaluate on the original validation dataset (5K images).
Out of 149,813 instances, 95,295 are detected by the face detector.
Detectron2 is run with commit: \href{https://github.com/facebookresearch/detectron2/tree/58e472e076a5d861fdcf773d9254a3664e045bf8}{58e472e076}

\paragraph{Cityscapes Instance Segmentation}
We train Mask R-CNN \cite{He2017} \href{https://github.com/facebookresearch/detectron2/blob/58e472e076a5d861fdcf773d9254a3664e045bf8/configs/Cityscapes/mask_rcnn_R_50_FPN.yaml}{R-50 FPN} using detectron2 \cite{wu2019detectron2} on the Cityscapes dataset \cite{cityscapes}.
The training dataset contains 2,975 images with 17,919 person instances (after filtration following \cref{sec:method}), and we evaluate on the original validation dataset (500 images).
Out of 17,919 instances, 4,456 were detected by the face detector.
Interestingly, this is a noticeably smaller percentage than for the COCO dataset, which we speculate is due to the dataset distribution (persons in COCO often face the camera, while they often do not in Cityscapes).

\paragraph{BDD100K Instance Segmentation}
We train Mask R-CNN \cite{He2017} \href{https://github.com/SysCV/bdd100k-models/blob/0935a8a/ins_seg/configs/ins_seg/mask_rcnn_r50_fpn_3x_ins_seg_bdd100k.py}{R-50 FPN} using MMDetection \cite{MMdetection} on the BDD100K dataset \cite{Yu2020}.
The training dataset contains 7K images with 9,954 person instances (after filtration following \cref{sec:method}), and we evaluate on the original validation dataset (1K images).
Out of 9,954 instances, 687 were detected by the face detector.
MMdetection is run with commit: \href{https://github.com/open-mmlab/mmdetection/tree/b95583270c57b3b0dc9c0523b2d1ebe46b755cca}{b95583270c}.

\begin{figure}
    \includegraphics[width=0.5\textwidth]{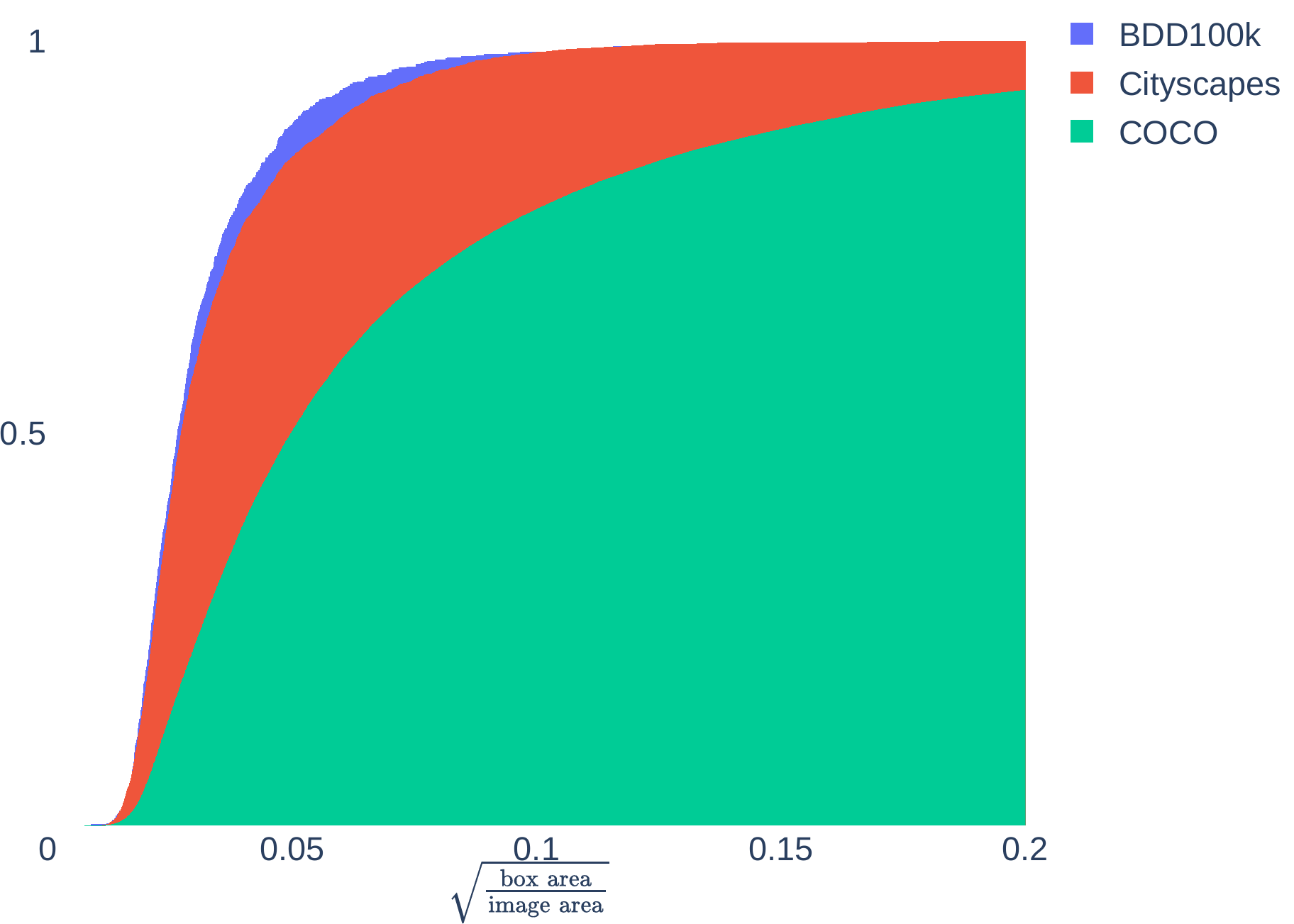}
    \caption{Cumulative histogram of average bounding box length normalized to image size.}
    \label{fig:bbox_area_histogram}
\end{figure}

\begin{table}

    \begin{center}
        \caption{
            Instance segmentation AP on the Cityscapes \cite{cityscapes} validation set with a Mask R-CNN \cite{He2017} \href{https://github.com/facebookresearch/detectron2/blob/main/configs/Cityscapes/mask_rcnn_R_50_FPN.yaml}{R-50 FPN}.
            \textbf{HM}=Histogram matching (\cref{sec:global_context}).
            \textbf{HM-LO}=Histogram matching via Latent Optimization (\cref{sec:global_context}).
            }
            \label{tab:cityscapes}
        \resizebox{\linewidth}{!}{
        \begin{tabular}{ccccc}
            \cmidrule[\heavyrulewidth]{2-5}
            
            & \textbf{Anonymization Method} & \textbf{AP} $\uparrow$ & \textbf{AP50} $\uparrow$ & \textbf{AP}$_\text{\textbf{person}}$ \\
            %\cline{2-5}
            & Original & $36.7 \pm 0.1$ $(\Delta)$  & $62.8 \pm 0.2$ & $35.0 \pm 0.2$  $(\Delta)$ \\
            \cmidrule{2-5}
            \parbox[t]{2mm}{\multirow{3}{*}{\rotatebox[origin=c]{90}{Face}}}& Blur & $36.4 \pm 0.2$ (-0.3) & $62.5 \pm 0.2$ (-0.3) & $34.9 \pm 0.1$ (-0.1) \\
            %\hline
            & Mask-out & $\mathbf{36.7} \pm 0.2$ (0.0) & $\mathbf{63.1} \pm 0.2$ (0.3) & $34.9 \pm 0.1$ (-0.1) \\
            %\hline
            & Realistic & $36.6 \pm 0.1$ (-0.1) & $62.8 \pm 0.3$ (0.0) & $\mathbf{35.0} \pm 0.1$ (0.0) \\
            \cmidrule{2-5}
            \parbox[t]{2mm}{\multirow{5}{*}{\rotatebox[origin=c]{90}{Body}}} & Blur & $31.4 \pm 0.2$ (-5.3) & $54.5 \pm 0.4$ (-8.3) & $2.1 \pm 0.1$ (-32.9) \\
            & Mask-out & $31.2 \pm 0.1$ (-5.5) & $53.2 \pm 0.1$ (-9.6) & $0.7 \pm 0.1$ (-34.3) \\
            & Realistic & $34.6 \pm 0.1$ (-2.1) & $59.0 \pm 0.3$ (-3.8) & $20.3 \pm 0.2$ (-14.7) \\
            & Realistic + HM & $34.3 \pm 0.2$ (-2.4) & $58.9 \pm 0.2$ (-3.9) & $21.3 \pm 0.3$ (-13.7) \\
            & Realistic + HM-LO & $\mathbf{34.8} \pm 0.2$ (-1.9) & $\mathbf{60.0} \pm 0.3$ (-2.8) & $\mathbf{21.5} \pm 0.1$ (-13.5) \\
            \cmidrule[\heavyrulewidth]{2-5}
        \end{tabular}}
        \end{center}
\end{table}

\begin{table}
    \begin{center}
        \caption{
        Instance segmentation AP on the BDD100K \cite{Yu2020} validation set with a Mask R-CNN \cite{He2017} \href{https://github.com/SysCV/bdd100k-models/blob/0935a8a3eb4c7442efdce2a8ce4b93fbe585be15/ins_seg/configs/ins_seg/mask_rcnn_r50_fpn_3x_ins_seg_bdd100k.py}{R-50 FPN}.
        }
        \label{tab:bdd100k}
        \resizebox{\linewidth}{!}{
        \begin{tabular}{ccccc}
            \cmidrule[\heavyrulewidth]{2-5}
            & \textbf{Anonymization Method} & \textbf{AP} $\uparrow$ & \textbf{AP50} $\uparrow$ & \textbf{AP}$_\text{\textbf{person}}$ \\
           
            & Original & $20.2 \pm 0.2 $ $(\Delta)$ & $34.9 \pm 0.4 $ $(\Delta)$ & $32.0 \pm 0.0 $ $(\Delta)$ \\
            \cmidrule{2-5}
            \parbox[t]{2mm}{\multirow{3}{*}{\rotatebox[origin=c]{90}{Face}}} & Blur & $20.5 \pm 0.1$ (0.3) & $35.9 \pm 0.1$ (1.0) & $31.7 \pm 0.1$ (-0.3) \\
            & Mask-out & $20.3 \pm 0.1$ (0.1) & $35.3 \pm 0.3$ (0.4) & $31.4 \pm 0.1$ (-0.6) \\
            & Realistic & $\mathbf{20.6} \pm 0.1$ (0.4) & $\mathbf{35.8} \pm 0.3$ (0.9) & $\mathbf{31.6} \pm 0.2$ (-0.4) \\
            \cmidrule{2-5}
            \parbox[t]{2mm}{\multirow{3}{*}{\rotatebox[origin=c]{90}{Body}}} & Blur & $15.4 \pm 0.1$ (-4.8) & $26.3 \pm 0.2$ (-8.6) & $0.5 \pm 0.0$ (-31.5) \\
            & Mask-out & $15.3 \pm 0.0$ (-4.9) & $25.5 \pm 0.1$ (-9.4) & $0.0 \pm 0.0$ (-32.0) \\
            & Realistic & $\mathbf{17.0} \pm 0.1$ (-3.2) & $\mathbf{28.9} \pm 0.4$ (-6.0) & $\mathbf{12.8} \pm 0.1$ (-19.2) \\
            \cmidrule[\heavyrulewidth]{2-5}
        \end{tabular}}
        \end{center}
        
\end{table}
\begin{table}
    \begin{center}
        \caption{
            Keypoint (Kp.) AP on the COCO \cite{Lin2014COCO} validation set with a Keypoint \href{https://github.com/facebookresearch/detectron2/blob/main/configs/COCO-Keypoints/keypoint_rcnn_R_50_FPN_3x.yaml}{R-50 FPN}  R-CNN  \cite{He2017}.
            }
            \label{tab:keypoint_COCO}
        \resizebox{\linewidth}{!}{
        \begin{tabular}{cccc}
            \cmidrule[\heavyrulewidth]{2-4}
             & \textbf{Anonymization Method} &\textbf{Box AP} $\uparrow$ & \textbf{Kp. AP} $\uparrow$ \\
             %\cline{2-4}
             & Original & $55.7 \pm 0.0$ $(\Delta)$  & $65.2 \pm 0.0$ $(\Delta)$  \\
            %\hhline{~===}
            \cmidrule{2-4}
            \parbox[t]{2mm}{\multirow{4}{*}{\rotatebox[origin=c]{90}{Face}}}&  Blur & $50.3 \pm 0.2$ (-5.4) & $53.5 \pm 0.2$ (-11.7) \\
            %\cline{2-4}
            & Mask-out & $49.9 \pm 0.2$ (-5.8) & $52.0 \pm 0.3$ (-13.2) \\
            %\cline{2-4}
            & Realistic & $54.3 \pm 0.1$ (-1.4) & $60.6 \pm 0.1$ (-4.6) \\
            & Realistic + HR Faces & $\mathbf{54.4} \pm 0.0$ (-1.3) & $\mathbf{60.8} \pm 0.2$ (-4.4) \\
            \cmidrule{2-4}
            \parbox[t]{2mm}{\multirow{3}{*}{\rotatebox[origin=c]{90}{Body}}} & Blur & $17.8 \pm 0.0$ (-37.9) & $4.4 \pm 0.1$ (-60.8) \\
            %\cline{2-4}
            & Mask-out & $17.4 \pm 0.1$ (-38.3) & $2.0 \pm 0.1$ (-63.2) \\
            %\cline{2-4}
            & Realistic & $\mathbf{24.0} \pm 0.1$ (-31.7) & $\mathbf{15.6} \pm 0.1$ (-49.6) \\
            \cmidrule[\heavyrulewidth]{2-4}
        \end{tabular}}
        \end{center}
\end{table}

\subsection{Effect of Face Anonymization}
We start our analysis by focusing on face anonymization.
On Cityscapes and BDD100k (\cref{tab:cityscapes}, \ref{tab:bdd100k}), we observe no significant performance difference from any type of face anonymization.
We note that realistic anonymization slightly outperforms mask-out anonymization for both datasets.
In \Cref{fig:bbox_area_histogram}, we find that the majority of boxes in BDD100K/Cityscapes cover less than 1\% of the image area.
Thus, it is not surprising that face anonymization has little impact on these datasets.

For COCO pose estimation (\cref{tab:keypoint_COCO}), face anonymization severely impacts performance, where both mask-out and blurring degrade keypoint AP by $>10\%$.
This performance drop is significant for bounding box AP as well, reflecting that the performance difference is not due to the inability to predict keypoints in the facial region.
Likely, this is due to learning that blurring/masking artifacts correlate to the human body.
Furthermore, we hypothesize that the major performance drop compared to Cityscapes and BDD100k is due to dataset distribution and not the task at hand.
To validate this, we train an instance segmentation model on the anonymized COCO datasets and observe a similar performance drop
\footnote{
    For mask-out, we observe a 6.7\% performance drop for Box AP for COCO instance segmentation, compared to a 10.4\% drop for Box AP for Keypoint R-CNN in \cref{tab:keypoint_COCO}.
    See \appExpDetailsCOCOInstance for more details.
}.

\paragraph{Refining COCO Faces}
Although realistic anonymization significantly improves over traditional methods, there remains a considerable degradation between it and the original COCO dataset.
We hypothesize that this degradation results from the following factors; limited synthesis quality, facial keypoint mismatch, and low-resolution synthesis.
As the generative model is not conditioned on facial keypoints, the synthesized identity will likely not match the annotated keypoints.
There exists keypoint guided anonymization models \cite{Maximov2020,Hukkelas2019,Sun2018}, which we leave for further work to investigate.
Furthermore, the generative model synthesizes faces at $128 \times 128$ resolution, introducing upsampling artifacts for any face above.
In total, we found 14,688 faces with an area larger than $128^2$.
To remove these upsampling artifacts, we employ a higher resolution ($256 \times 256$) face synthesis model from DeepPrivacy2 \cite{Hukkelas2022a} to anonymize any face larger than $128\times 128$.
This slightly improved downstream use (marked \emph{Realistic + HR Faces} in \cref{tab:keypoint_COCO}), supporting our hypothesis that upsampling artifacts can degrade image utility for COCO keypoint detection training.

\subsection{Effect of Full-Body Anonymization}

For full-body anonymization, we observe a substantial decline in performance for both traditional and realistic anonymization methods (\cref{tab:cityscapes}, \ref{tab:bdd100k}, \ref{tab:keypoint_COCO}).
Traditional anonymization leads to a complete degradation in performance, whereas realistic anonymization improves this significantly.
Interestingly, the performance of realistic full-body anonymization on BDD100K \cite{Yu2020} is noticeably worse than for Cityscapes \cite{cityscapes}, which we discuss further below.

Clearly, realistic full-body anonymization significantly degrades the performance compared to the original dataset, which we attribute to the following three issues: keypoint detection errors, synthesis limitations, and global context mismatch.
Synthesizing realistic human bodies is difficult, and current models may introduce severe visual artifacts for many contexts.
Furthermore, current methods rely on a crop-based anonymization method (discussed in \Cref{sec:global_context}), which can result in synthesized identities that do not fit the global context of the image.
\Cref{sec:global_context} introduced naive histogram matching and HM-LO to mitigate this issue, which we find to significantly improve results on the Cityscapes dataset (\Cref{tab:cityscapes}).

\paragraph{BDD100k \vs Cityscapes}
The decline in performance is significantly more prominent for BDD100k than Cityscapes, despite both datasets being collected for the same purpose.
We suspect this discrepancy stems from two sources; keypoint annotations and dataset resolution.
First, ViTPose \cite{Xu22VITPose} detects keypoints for 95.8\% of the instances in the Cityscapes dataset, whereas it only detects for 85.5\%  in the BDD100k dataset.
Secondly, the BDD100k images are of lower resolution (720p) than Cityscapes ($2048 \times 1024$).
This results in 36\% of the instance crops having an area $<32^2$, compared to 24\% for Cityscapes.
While lower-resolution bodies are easier to synthesize in theory, the employed generative model operates at the resolution $288 \times 160$, and major deviations from this resolution can induce visual artifacts.
For example, if we do not anonymize any detections $<32^2$, BDD100k AP$_{\text{person}}$ is increased from 12.8\% to 19.9\%.
In contrast, this increases AP$_{\text{person}}$ from 20.3\% to 23.4\% for Cityscapes.

\subsection{Ablations}
\paragraph{Do Larger Models Generalize Better?}
\begin{figure}
    \includegraphics[width=.5\linewidth]{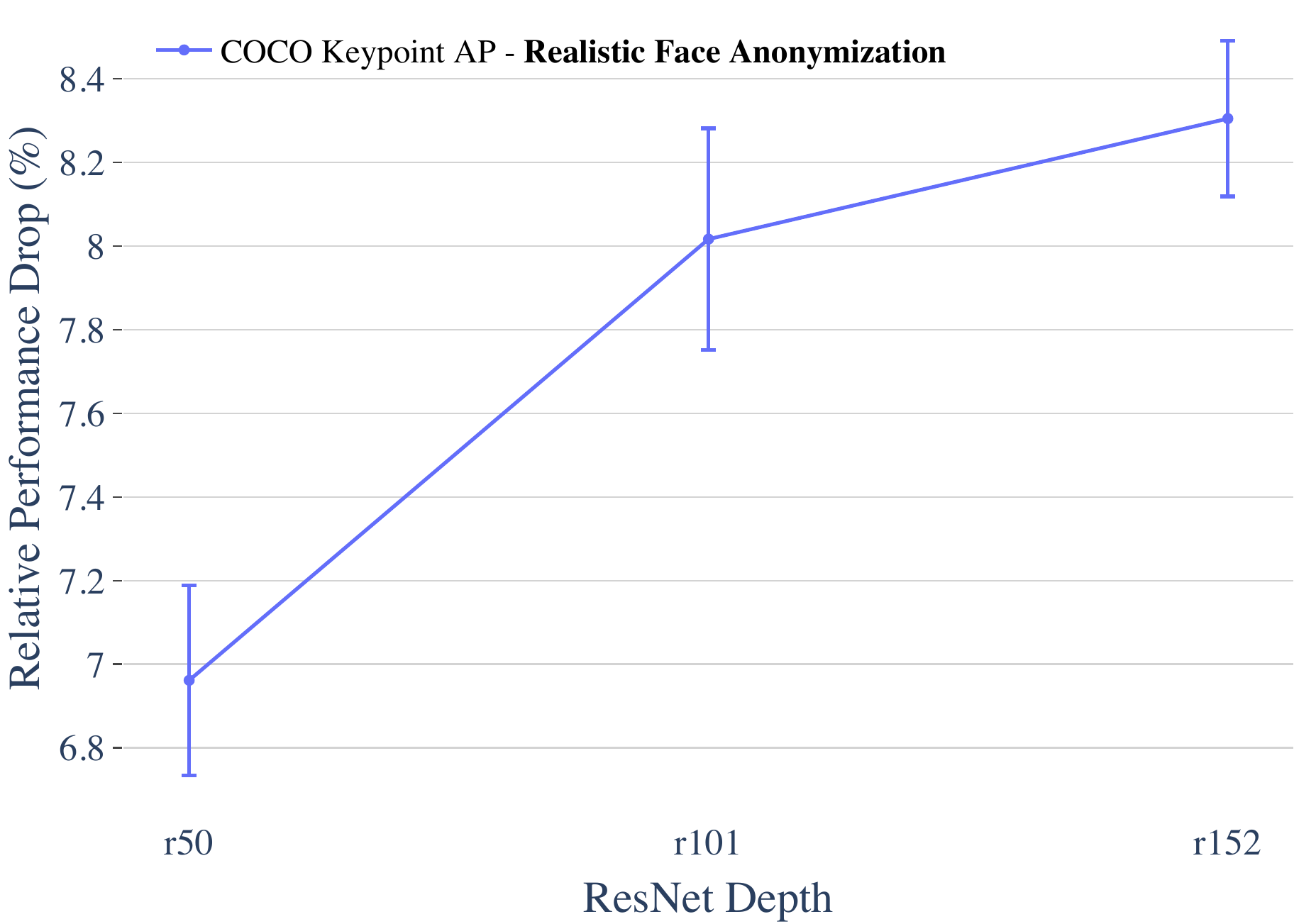}%
    \includegraphics[width=.5\linewidth]{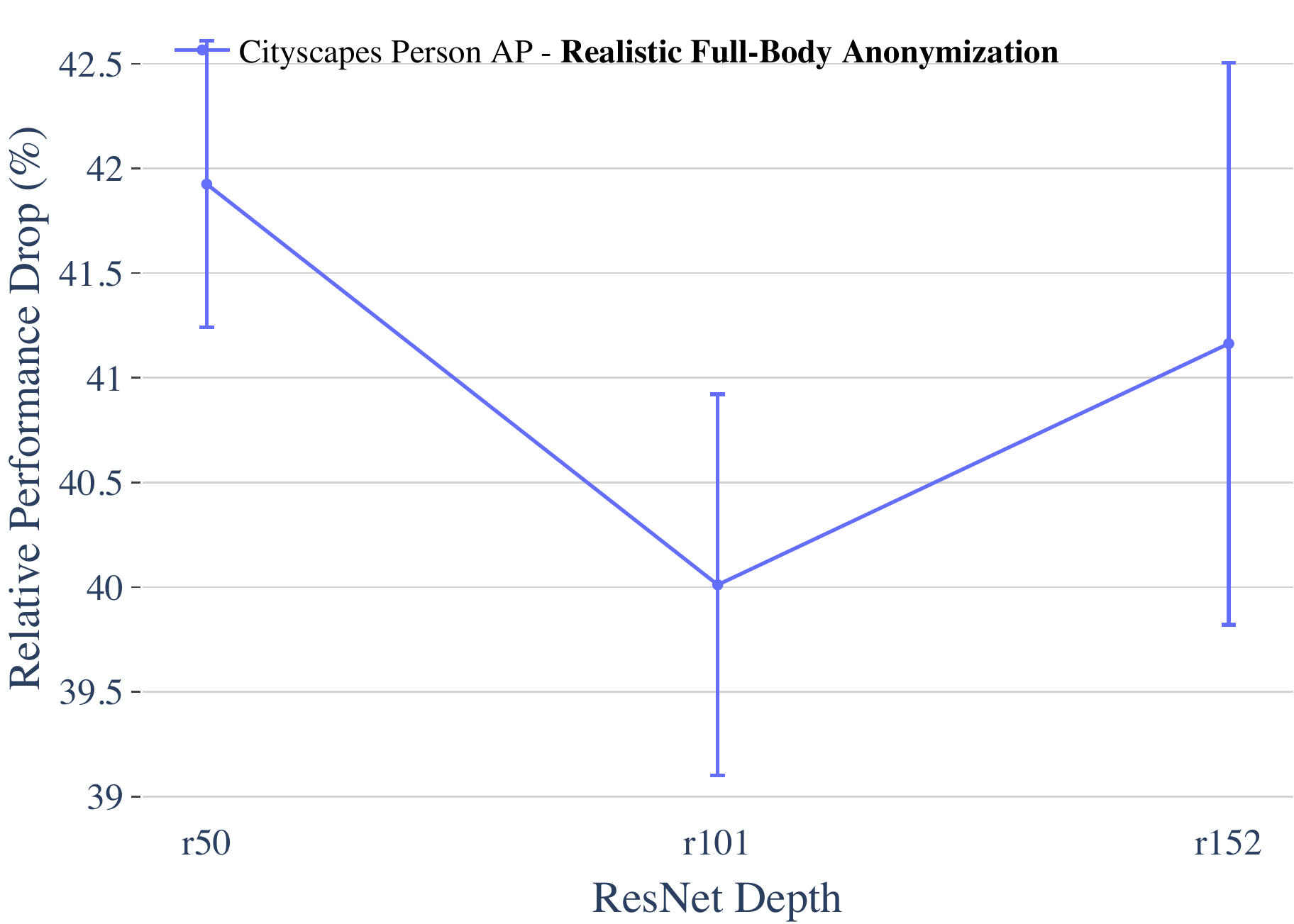}
    \caption{The relative performance drop of realistic anonymization (face or body) for different ResNet depths.}
    \label{fig:larger_models}
\end{figure}
Zhou \etal \cite{Zhou2022} observes that deeper models are less impacted by realistic image anonymization.
In our experiments, we observed the reverse to be true.
We train a ResNet-50, 101, and 152 and compare the relative performance drop of realistic anonymization compared to the original dataset.
We investigate this for realistic face anonymization on COCO and full-body anonymization for Cityscapes.
\Cref{fig:larger_models} reflects that larger models perform worse for both the COCO, whereas it is not clear for the Cityscapes dataset.

\paragraph{Diversity vs. Quality Trade-off}
\begin{table}

    \begin{center}
        \caption{
            Instance segmentation AP on the Cityscapes \cite{cityscapes} validation set with full-body anonymization using different latent sampling strategies.
            Results from Mask R-CNN \cite{He2017} \href{https://github.com/facebookresearch/detectron2/blob/main/configs/Cityscapes/mask_rcnn_R_50_FPN.yaml}{R-50 FPN}.
            }
        \label{tab:cityscapes_truncation}
        \resizebox{\linewidth}{!}{
        \begin{tabular}{cccc}
            \toprule
             \textbf{Anonymization Method} & \textbf{AP} $\uparrow$ & \textbf{AP50} $\uparrow$ & \textbf{AP}$_\text{\textbf{person}}$ \\
            %\cline{2-5}
            
            Original & $36.7 \pm 0.1$ $(\Delta)$  & $62.8 \pm 0.2$ $(\Delta)$  & $35.0 \pm 0.2$ $(\Delta)$ \\
            \midrule
            No Truncation & $34.0 \pm 0.2$ (-2.7) & $57.7 \pm 0.5$ (-5.1) & $18.6 \pm 0.2$ (-16.4) \\
            Unimodal Truncation & $33.9 \pm 0.2$ (-2.8) & $58.1 \pm 0.3$ (-4.7) & $19.7 \pm 0.5$ (-15.3) \\
            Multi-modal Truncation (\textbf{Default}) & $\mathbf{34.6} \pm 0.1$ (-2.1) & $\mathbf{59.0} \pm 0.3$ (-3.8) & $\mathbf{20.3} \pm 0.2$ (-14.7) \\
            \bottomrule
        \end{tabular}}
        \end{center}
\end{table}

GANs can trade off the diversity of samples with quality through the truncation trick \cite{Brock2018}.
Specifically, by interpolating the input latent variable $z \sim \mathcal{N}(0,1)$ towards the mode of $\mathcal{N}(0,1)$, generated diversity is traded off for improved quality.
This leaves the question, what is best for anonymization purposes?
Limited diversity might result in a detector primarily being able to detect a small diversity of the population, whereas limited quality might reduce transferability to real-world data.

We explore the use of the truncation trick for anonymization purposes, where we investigate the use of no truncation, multi-modal truncation \cite{Mokady2022} \footnote{
    Multi-modal truncation \cite{Mokady2022} approximates multiple modes of the latent distribution, enabling sampling high-quality images while minimizing the loss of diversity.
    We estimate 512 cluster centers following \cite{Hukkelas2022a}.
}, and standard truncation \cite{Brock2018}.
Note that in all other experiments, multi-modal truncation is used for full-body anonymization, while we use no truncation for face anonymization.

\Cref{tab:cityscapes_truncation} reflects that both standard and multi-modal truncation performs substantially better than no truncation for $AP_{\text{person}}$.
Furthermore, we observe that multi-modal truncation further improves over standard truncation.

\paragraph{Does Anonymization Impact Other Classes?}
For many tasks, person detection is not the intended task of the anonymized data (\eg road damage detection \cite{RDD2022}).
Thus, we investigate the impact of anonymization where person detection is not part of the task.
To answer this, we re-train the instance segmentation for the Cityscapes dataset and exclude the "person" class from the segmentation task.

Our experiment (see \appExpDetailsCityscapesNoPersons) reflects that full-body anonymization  does not impact the detection of the following classes: bus, car, motorcycle, train, or truck.
However, we do notice a performance drop for detecting  "rider" and "bicycle".
We believe this is due to detection overlaps.

\section{Conclusion}
\label{sec:discussion}
In this work, we investigated the impact of anonymization for training computer vision models, with a focus on autonomous vehicle datasets.
Our experiments reflect that face anonymization (obfuscation and realistic) has little to no impact for instance segmentation on the BDD100K \cite{Yu2020} and Cityscapes \cite{cityscapes} datasets.
In contrast, face obfuscation severely degrades the performance of keypoint detection models on the COCO \cite{Lin2014COCO} dataset, as faces are more prevalent in comparison to the BDD100k and Cityscapes datasets.
We find that realistic face anonymization can significantly reduce this performance drop.
Furthermore, we find that full-body obfuscation severely impairs performance on all datasets, where realistic full-body anonymization can notably  alleviate this issue.
In summary, our findings reflect that realistic anonymization is a superior option compared to traditional methods.
However, they are not a complete substitute for real data, especially for full-body anonymization, as current generative models can often produce unnatural humans that do not fit the given context.

\paragraph{Societal Impact}
Computer vision models are becoming increasingly adopted for solving challenging tasks everywhere in our society, from manufacturing to driving our cars.
These models require task-specific training data to specialize for the task at hand.
Collecting such data is troublesome due to privacy legislation, especially for autonomous vehicles which operate in environments where individuals appear everywhere.
Our findings indicate that realistic anonymization can effectively substitute the original data, encouraging companies to protect individuals' privacy without compromising model performance.
Our main societal concern is that we do not advocate that the anonymization methods studied in this paper give any sort of privacy guarantee.
The detailed discussion in \Cref{sec:related_work} clarifies that face anonymization and image blurring are questionable with respect to privacy.
Furthermore, anonymized bodies could still be identified, \eg  from gait recognition \cite{Abarca2021handbook}.

\subsection{Limitations and Further Work}
\paragraph{Limitations}
The primary limitation of our study is the reliance on automatic annotations, where we use DSFD \cite{Li2018c} for face detections, and ViTPose \cite{Xu22VITPose} for keypoint annotations.
While the performance of these methods is impressive, they introduce ambiguity in our results, questioning if the current performance degradation is due to annotation errors or synthesis limitations.
Furthermore, due to the filtering criteria for full-body anonymization and automatic annotation of faces, we are not able to anonymize all individuals in the images.
Finally, it is also worth mentioning that our analysis is restricted to ResNet \cite{He2016resnet} and R-CNN \cite{Ren2015} based models and that other architectures (\eg YOLO \cite{Bochkovskiy2020}) may respond differently to anonymization artifacts.

\paragraph{Further Work}
Our explorative analysis of current realistic anonymization techniques highlights several areas of improvement and limitations.
To the best of our knowledge, all current anonymization techniques rely on a crop-based anonymization method to improve synthesis quality.
However, this can result in a mismatch between the synthesized identity and the global image.
For example, the synthesized identity may not align with the global context of the image despite fitting the local crop given to the generative model.
To mitigate this, we show that histogram equalization can reduce the impact of this, but we note that histogram equalization is far from the optimal solution.
Furthermore, our experiments reflect that there are major practical difficulties remaining in effectively employing generative models for anonymization.
For example, current anonymization techniques operate at a fixed synthesis resolution, where large deviations from the operating resolution (\eg bodies smaller than $32^2$) result in unnatural images, which impacts performance.
Finally, we note that there are several intriguing and unexplored challenges to handle for synthesizing human figures for anonymization in autonomous vehicles.
\Eg handling multi-view consistency, temporal consistency, or ensuring that the synthesized demography matches the demography of the original data.

\paragraph{Acknowledgement}
The computations were performed on resources provided by the NTNU IDUN/EPIC computing cluster \cite{sjalander2019epic}.
Furthermore, we thank Rudolf Mester for his general support and thoughtful discussion.
%%%%%%%%% REFERENCES
{\small
\bibliographystyle{ieee_fullname}
\bibliography{main.bib}
}

\setcounter{section}{0}
\renewcommand{\thesection}{\Alph{section}}

\end{document}